# Gaussian Process Regression – Neural Network Hybrid with Optimized Redundant Coordinates

Sergei Manzhos[1], Manabu Ihara

School of Materials and Chemical Technology, Institute of Science Tokyo, Ookayama 2-12-1, Meguro-ku, Tokyo 152-8552 Japan

**Abstract**

Recently, a Gaussian Process Regression – neural network (GPRNN) hybrid machine learning method was proposed, which is based on additive-kernel GPR in redundant coordinates constructed by rules [J. Phys. Chem. A 127 (2023) 7823]. The method combined the expressive power of an NN with the robustness of linear regression, in particular, with respect to overfitting when the number of neurons is increased beyond optimal. We introduce opt-GPRNN, in which the redundant coordinates of GPRNN are optimized with a Monte Carlo algorithm and show that when combined with optimization of redundant coordinates, GPRNN attains the lowest test set error with much fewer terms / neurons and retains the advantage of avoiding overfitting when the number of neurons is increased beyond optimal value. The method, opt-GPRNN possesses an expressive power closer to that of a multilayer NN and could obviate the need for deep NNs in some applications. With optimized redundant coordinates, a dimensionality reduction regime is also possible. Examples of application to machine learning an interatomic potential and materials informatics are given.

---

[1] Author to whom correspondence should be addressed. E-mail: Manzhos.s.aa@m.tiech.ac.jp, Tel. & Fax: +81-3-5734-3918

# 1 Introduction

The popularity of machine learning (ML) in various applications has much to do with the fact that common ML methods such as neural networks (NN) [1,2], kernel regressions (KR) [3,4] or decision trees (DT) [5–7] are general – the same method and code can be applied to data from social sciences just as well as they can be in quantum chemistry or physics or materials informatics. This generality and versatility come at a cost of being more data-intensive than models (analytic formula) based on domain knowledge. While simple linear regressions have limited expressive power, NN, KR, and DT can in principle represent any input-output relation expressible by smooth functions to any degree of accuracy, given enough data. The accuracy in all relevant space, typically exemplified by prediction errors on a test set, is then limited by the density of sampling and featurization and not by the expressive power of the method. When hyperparameters (numbers of layers and neurons of an NN, length and noise parameters of a KR, depth of a DT, etc.) are optimal, all these method often can achieve similar levels of test set error (so much so that if they do not, it is an indication of non-optimal hyperparameters) [8]. The finiteness of available training data creates the problem of overfitting. In particular in the case of NNs, one has to identify an optimal architecture, and the test set error increases if the number of neurons is increased beyond optimal.

We note that both NNs and KRs can be viewed as linear regressions over nonlinear basis sets. A kernel regression of a function $f(x), x \in R^D$ is a linear regression over a non-linear basis made from kernel functions $k(x, x')$,

$$f(x) = \sum_{i=1}^{M} c_i k(\xi \,|\, x, x^{(i)})$$

(1)

where $x^{(i)}, i = 1, \dots, M$ are training data and $\xi$ is the vector of hyperparameters (such as the width parameter of an RBF kernel, the order of the polynomial kernel, etc.). KR is typically regularized, to make the inversion of the covariance matrix $K$ (with elements $K_{ij} =$ $k(\boldsymbol{x}^{(i)}, \boldsymbol{x}^{(i)})$) stable; different ways to regularize give rise to different flavors of KR such as GPR (Gaussian process regression) or SVR (support vector regression) [3,8]. In the following, we will use GPR.

An NN can also be viewed as linear regression over a nonlinear basis. This is obvious in the case of a single-hidden later NN (which is already a universal approximator [9–22] with any smooth nonlinear neuron activation function [23]):

$$f(\boldsymbol{x}) = \sigma_{out}\left(\sum_{i=1}^{N} c_i \sigma(\boldsymbol{w}_i \boldsymbol{x} + b_i) + b_{out}\right)$$

(2)

where $N$ is the number of neurons (a hyperparameter), $\sigma$ are hidden layer neuron activation functions (NAF) that are typically the same for all neurons but do not have to be, $b_i$ are hidden layer biases, are $\boldsymbol{w}_i$ weights, and $\sigma_{out}, b_{out}$ are output NAF and bias, respectively. The output neuron and bias can be omitted (notionally moved to the lefthand side) without loss of generality, and one is then left with a linear regression over a basis of $\sigma_i(\boldsymbol{x}) = \sigma(\boldsymbol{w}_i|\boldsymbol{x})$. A multilayer NN can also be viewed as a linear expansion over a basis made from the NAFs of the last hidden layer:

$$f(\boldsymbol{x}) = \sum_{k_n=0}^{N_n} w_{l,k_n}^{(n)} \sigma_{n,k_n}\left(\sum_{k_{n-1}=0}^{N_{n-1}} w_{k_n,k_{n-1}}^{(n-1)} \sigma_{n-1,k_{n-1}}\left(\cdots \sum_{k_1=0}^{N_1} w_{k_2 k_1}^{(2)} \sigma_{1,k_1}\left(\sum_{i=0}^{d} w_{k_1 i}^{(1)} x_i\right)\right)\right)$$
$$\equiv \sum_{i=0}^{N_n} c_i\, \theta_i(\boldsymbol{W}|\boldsymbol{x})$$

(3)

where we omitted output NAF and for brevity subsumed the biases into the weights by defining a dummy variable $x_0 \equiv 1$. We also kept subscripts on $\sigma$ to highlight the fact that they need not be the same neither from layer to layer nor within the same layer. The effective basis $\theta_i(\boldsymbol{W}|\boldsymbol{x}) = \sigma_{n,i}\left(\sum_{k_{n-1}=0}^{N_{n-1}} w_{k_n,k_{n-1}}^{(n-1)} \sigma_{n-1,k_{n-1}}\left(\ldots \sum_{k_1=0}^{N_1} w_{k_2 k_1}^{(2)} \sigma_{1,k_1}\left(\sum_{i=0}^{d} w_{k_1 i}^{(1)} x_i\right)\right)\right)$ is parameterized by the weights $\boldsymbol{W}$ of all previous layers.

In KR, Eq. (1), the hyperparameters are the same for all basis functions – kernels, the regression is linear. In an NN, the parameters of each basis function are different, they are nonlinear regression parameters. This gives an NN a higher expressive power; in particular, the basis functions – neurons of the last hidden layer can have any shape if the NN is deep enough, even though individual NAFs are simple-shaped functions such as sigmoid or exponents, etc. [24,25]. This flexibility comes at a cost of many non-linear optimization parameters; for which reason, when the data are sparse and dangers of overfitting are severe, KR can outperform NN [26]. Large NNs also incur a higher CPU cost of training and recall. KR can be made very robust in the sense of achieving reliable ML from very sparse data by using additive kernels [27] (kernels in the form of expansion over orders of coupling [28–32] resulting in $f(\boldsymbol{x})$ expressed as such expansion [30,33–36]). In the limiting case of simple additive models when $k(\boldsymbol{x},\boldsymbol{x}') = \sum_{d=1}^{D} k_i(x_i, x_i')$ and $f(\boldsymbol{x}) \approx \sum_{d=1}^{D} f_i(x_i)$, the terms $f_i(x_i)$ can be well-defined even from very few data [37] and do not suffer from issues associated with (non-additive) high-dimensional kernels [38]. Additive kernel models by construction cannot recover inter-feature coupling terms of higher orders than the order of the additive kernel; however, higher-order terms may be unrecoverable in principle due to data sparsity [34,39].

In Ref. [40], we proposed a method, called GPRNN, that combined key advantages of KR and NN, namely, robustness associated with the absence of nonlinear regression in KR and expressive power of NN. We recognized that Eq. (2) (once output neuron and bias are omitted) is a simple additive model in redundant coordinates $\boldsymbol{y} = \boldsymbol{W}\boldsymbol{x}, \boldsymbol{y} \in R^N$, where $\boldsymbol{W}$ is the matrix of weights with rows $\boldsymbol{w}_i$:

$$f(\boldsymbol{x}) = f(\boldsymbol{y}(\boldsymbol{x})) = \sum_{i=1}^{N} f_n(y_n) = \sum_{i=1}^{N}\sum_{k=1}^{M} a_{nk} k\left(y_n, y_n^{(k)}\right)$$

(4)

Here each univariate component function $f_n(y_n)$ of the additive model in $\boldsymbol{y}$ is constructed with GPR using a one-dimensional kernel $k(y_n, y'_n)$. All terms $f_n(y_n)$ can be constructed simultaneously in one linear step by defining a first-order additive kernel in $\boldsymbol{y}$, $k(\boldsymbol{y}, \boldsymbol{y}') = \sum_{i=1}^{N} k(y_n, y'_n)$, and performing GPR in $\boldsymbol{y}$. $k(y_n, y'_n)$ can in general depend on $n$, but in practice, the same general kernel (such as a Matern kernel, a polynomial kernel, etc.) can be used for all $n$. The functions $f_n(y_n)$ are then optimal in the least squares sense for given data and $\boldsymbol{W}$. $\boldsymbol{W}$ were not fitting parameters but were defined by rules such as taking $\boldsymbol{w_i}$ as elements of an $N$-dimensional pseudorandom sequence [41]. In the $\boldsymbol{x}$ space (the original feature space), Eq. (4) is equivalent to a single hidden later NN with optimal NAF for each neuron, while in the $\boldsymbol{y}$ space and algorithmically, it is additive GPR. The absence of nonlinear optimization resulted in remarkable stability with respect to overfitting when $N$ is increased beyond optimal (see e.g. Figs. 7 and 9 of and Fig. 3 of Ref. [42]). Avoiding high-dimensional kernels also makes the method robust with respect to sparse data [8,38,43,44]. As magnitudes of $f_n(y_n)$ can serve to gauge the importance of features and their combinations, and the method, while being general, can generate insight [45,46]. The method is also amenable to symbolic regression [47–51] to produce ML-guided analytical formulas that are advantageous for interpretability, portability, and computation speed [52].

Absence of optimization of $\boldsymbol{W}$ also has disadvantages. While in the absence of optimization, increasing $N$ is inexpensive (the only non-trivial extra cost vs. conventional, non-additive, GPR is the summation in the additive kernel), the result is a larger number of terms / neurons needed to achieve a similar accuracy to that of an optimal NN, see Ref. [53] for a comparison. When used for symbolic regression, more terms would result in a more complicated formula. Importantly, as is clear by comparing Eqs. (3) and (4), the optimization of $\boldsymbol{W}$ in GPRNN would result in the same expressive power as that of a deep NN, in principle allowing to obviate the need for a multilayer NNs and associated CPU costs and issues related to nonlinear optimization of a large number of parameters in deep NNs. Optimization of $\boldsymbol{W}$ of Eq. (4) is equivalent to the optimization of weights of a single hidden layer NN, while the optimal shapes of $f_n(y_n)$ constructed with additive GPR further increase expressive power.

In this work, we therefore explore GPRNN with optimized $W$ (opt-GPRNN) and consider the effect of optimization on accuracy and on the robustness of the method with respect to overfitting when $N$ is increased beyond optimal. Using ML force fields and materials informatics as example applications, we show that when combined with optimization of $W$ i.e. of redundant coordinates $y$, GPRNN retains the advantage of suppressing overfitting when the number of neurons is increased beyond optimal while resulting in fewer neurons needed to achieve the lowest test set error. With optimized redundant coordinates, dimensionality reduction regime also becomes possible.

## 2  Methods

The opt-GPRNN calculations were done in MATLAB (R2024b). GPRNN of Eq. (4) is easy to code in any ML programming environment: one defines $W$ and therefore $y$, one defines the additive kernel , $k(\mathbf{y}, \mathbf{y}') = \sum_{i=1}^{N} k(y_n, y_n')$ with the desired $k(y_n, y_n')$, and one does GPR in $y$ with any GPR engine. A MATLAB code is available (see Data availability statement). We take rows of $W$ as elements of a $N$-dimensional Sobol sequence, and we also include the original $D$ coordinates into $y$, i.e. $\mathbf{y}^{(i)} = \mathbf{x}^{(i)}$ for $i = 1, …, D$, and $\mathbf{y}^{(i)} = \mathbf{s}_{i-D+1}$ for $i = D+1$, …, $N$, where $\mathbf{s}_i$ are elements of a $N$-dimensional Sobol sequence (we skip the first element which is a null vector). In this work, we use an isotropic RBF kernel $k(y_n, y_n') = \exp((y_n - y_n')^2/2l^2)$. The features are scaled on the unit cube.

The optimization of $W$ is done with a Monte Carlo approach: at one optimization step, each $\mathbf{w}_i$ is perturbed as $\mathbf{w}_i = \mathbf{w}_i + rRc$, where is a random vector of unit length, $R$ is a random number on [0, 1], and $c$ is a coefficient. The results are not very sensitive to the choice of $c$, we used $c$ = 0.05. $W$ is updated if the training or validation set error decreases. We use a Monte Carlo approach as it does not require more complex coding of kernel-type dependent analytic derivatives and can easily be used with any kernel, while numeric derivatives are unattractive for high-dimensional ($N×D$-dimensional) optimization. With optimized $W$, it becomes possible to have a dimensionality reduction regime when $N < D$. In this case, the original $D$ coordinates are not included in $y$, and $\mathbf{y}^{(i)} = \mathbf{s}_{i+1}, i = 1, …, N$. The maximum number of steps was set to 2000.

With the optimization of redundant coordinates, the method in principle possesses an expressive power closer to that of a multilayer NN and could obviate the need for deep NNs in some applications. While $\theta_i(W|x)$ of the deep NN of Eq. (3) are fixed-shape NAFs whose arguments are complex non-linear functions of $x$, the fact that $f_n(y_n)$ can have arbitrary shapes partially compensates for the fact that the neuron argument $y_n = Wx$ is restricted to a linear form in $x$. The partial equivalence is easy to see from the Taylor expansions of $\theta_i(W|x)$ and $f_n(y_n) = f_n(w_n x)$ around the same $x_0$,

$$\theta_i(x) = \theta_i(x_0) + \nabla \theta_i(x_0) \Delta x + \frac{1}{2} (\Delta x)^T H \, \Delta x + \cdots$$

$$f_n(y_n(x)) = f_n(y_{n0}) + \left.\frac{df_n}{dy_n}\right|_{y_{n0}} \Delta y_n + \frac{1}{2} \left.\frac{d^2 f_n}{dy_n^2}\right|_{y_{n0}} (\Delta y_n)^2 + \cdots$$

$$= f_n(y_n(x_0)) + \left(w_n \left.\frac{df_n}{dy_n}\right|_{y_{n0}}\right) \Delta x + \frac{1}{2} (\Delta x)^T \left(w_n w_n^T \left.\frac{d^2 f_n}{dy_n^2}\right|_{y_{n0}}\right) \Delta x + \cdots$$

(5)

where $y_{n0} = w_n x_0$, and $H$ is the Hessian matrix ($H_{ij} = \frac{\partial^2}{\partial x_i \partial x_j} \theta_i(x)$). These expansions can be only approximately matched per term in each order of the resulting polynomial. Optimal $N$ with opt-GPRNN is therefore expected to be lower than in GPRNN but somewhat higher than that of the *best* multilayer NN (this is exactly what we observe in section 3.1 below).

Comparisons to conventional neural networks of different sizes and depths were done in MATLAB using hyperbolic tangent NAF and Levenberg-Marquard optimization algorithm, which provided the best results. The maximum number of epochs was set to 2000.

We test the method on two applications. One is machine learning the potential energy surface of $H_2O$ molecule in the spectroscopically relevant region up to 20,000 cm$^{-1}$ above the equilibrium geometry. The dataset comprises three Radau coordinates (the features) and the value of potential energy vs. equilibrium geometry (the target). The same dataset was used to test the original GPRNN before [40,54] and was also used with other ML methods [35,40,55] (the data are also available for download from those references), so that the relative performance of the present method can easily be judged. The desired level of ML accuracy (test set error) for a spectroscopically accurate small molecule PES is on the order of 1 cm$^{-1}$. It is achievable with about 1000 training points. We use 1000 randomly selected points for training and 3000 for testing. That is, this example tests the method in a high accuracy regime and with an abundant test set to reliably gauge the global accuracy of the fit. The kernel length parameter and noise parameter for this dataset, unless stated otherwise, were 0.5 and $1 \times 10^{-13}$, respectively, selected by grid search.

The second example application is from the field of materials informatics: we machine-learn zero-point vibrational energies (ZPE) of molecules from the QM9 [56] dataset containing molecules with C, O, N, F, and H atoms. The descriptors were the eigenspectrum of the Coulomb's Matrix (ECM) [57] generated from the atomic coordinates of the molecules. A subset of QM9 molecules is used containing 16 atoms, with the dimensionality of the feature space $D$ = 16 for all molecules, which resulted in 14,252 structures. We used 3000 structures for training and 10,000 structures for testing. The kernel length parameter and noise parameter for this dataset were 0.5 and $1 \times 10^{-5}$, respectively.

Both examples were selected so that large test sets could be used to reliably gauge the performance of the method. The random seed was fixed when doing random training-test set splits, to focus on the effect of *W* optimization and trends with *N* and on comparison between the methods.

## 3   Results

*3.1   The interatomic potential of the water molecule*

In Figure 1, top panel, we show training and test rmse values for different *N*, with and without *W* optimization, when machine learning the interatomic potential of the water molecule. Optimization allows obtaining the best possible test set error with a much smaller number of terms *N* (about 30 instead of about 60). The results without optimization follow those on Ref. [40]. Remarkably, the test set error does not increase when *N* is increased beyond its optimal value also with optimization, it follows a plateau at about 0.3 cm$^{-1}$. One might wonder how this is possible. This is because at any *W*, the calculation that is performed is additive GPR in the *y* space, and its overfitting performance is governed by kernel hyperparameters rather than *N*. To illustrate this, in Figure 1, bottom panel we also show the results with suboptimal kernel lengths spanning values differing by a factor of about 3. Overfitting with *N* increasing beyond its optimal value is avoided with each, but the test set error plateaus at a higher value of up to about 1.3 cm$^{-1}$ (which still results in a spectroscopically accurate potential).

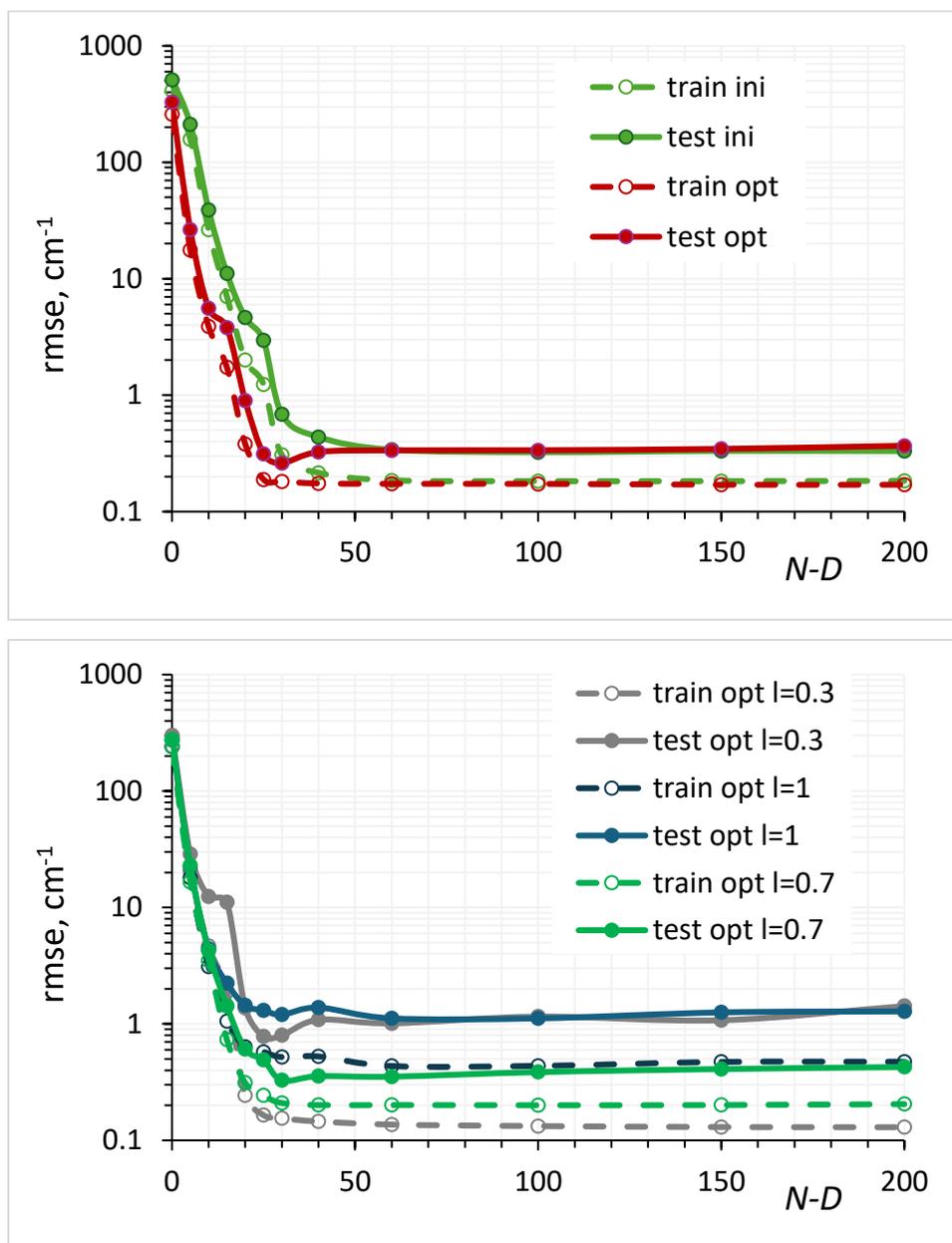

Figure 1. Top panel: the behavior of training and test set errors when fitting the interatomic potential of the water molecule with GPRNN ("ini") and opt-GPRNN ("opt") as a function of the number of terms, at an optimal kernel length parameter ($l = 0.5$). The abscissa shows the number of redundant coordinates $N - D$. Bottom panel: the behavior of training and test set errors with opt-GPRNN when using sub-optimal kernel length parameters ($l = 0.3, 0.7, 1.0$). Note the logarithmic scale.

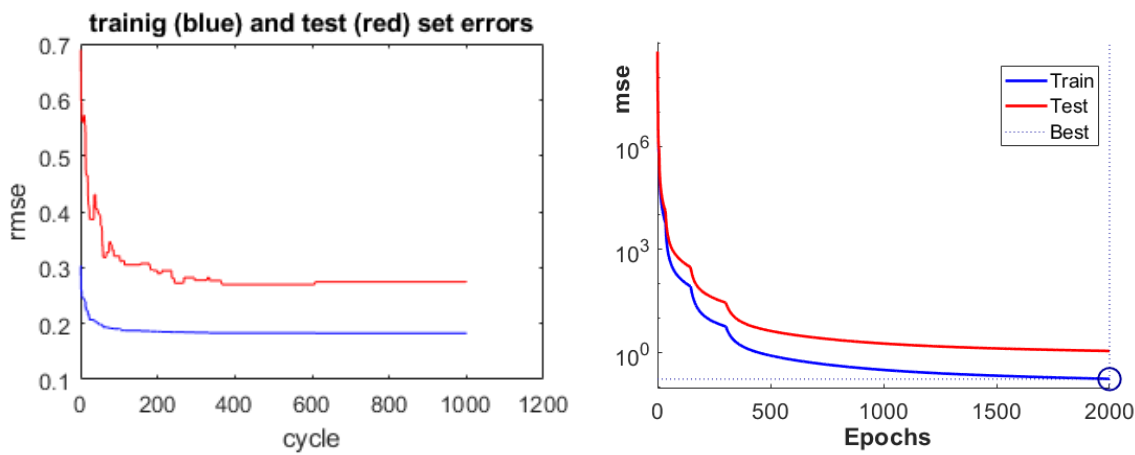

Figure 2. Learning curves of opt-GPRNN (*N* = 33) (left) and best conventional NN (right). As errors at the start of training are high with an NN, they have to be shown on the logarithmic scale. GPRNN errors before optimization are substantially lower that those of the NN before training due to the optimal nature of the component functions.

We searched the best possible conventional NN by scanning the numbers of layers and neurons per layer, performing 5 runs with each NN size to account for local minima. The best NN we found had 4 layers with 10 neurons per layer, achieving a training set error of 0.29 cm$^{-1}$ (average over 5 runs ranging 0.19-0.61) and the test set error was 0.58 cm$^{-1}$ (average over 5 runs ranging 0.31-1.02). The best single hidden layer NN had 50 neurons, achieving a training set error of 0.30 cm$^{-1}$ (average over 5 runs ranging 0.26-0.36) and the test set error was 1.16 cm$^{-1}$ (average over 5 runs ranging 0.70-1.57). The conventional NN required many epochs to achieve high accuracy. For example, when limiting the number of epochs to 1000, the best NN still had 4 hidden layers with 10 neurons per layer with a training set error of 0.35 cm$^{-1}$ (average over 5 runs ranging 0.29-0.44) and the test set error was 1.17 cm$^{-1}$ (average over 5 runs ranging 0.81-1.57). The best single hidden later NN trained over 1000 epochs had 50 neurons and average training/test set errors of 0.53/2.01 cm$^{-1}$ (ranging over 5 runs (0.39-0.67)/(1.30-2.86), respectively). Note that we used the LM algorithm that converged the fastest. In Figure 2, we show examples of learning curves for optimal opt-GPRNN ($N = 33$) and best conventional NN. We see that despite the fact that the MC algorithm makes many steps not resulting in $W$ updates, the total number of steps to achieve errors near the best possible errors is lower that of a conventional NN (that uses analytic derivatives and identical neurons to speed up calculations). Errors at the start of training are very high with an NN (which is why they have to be shown on the logarithmic scale. GPRNN errors before optimization are substantially lower than those of an NN before training, as neuron shapes are optimal by virtue of additive GPR.

With a conventional NN, overfitting quickly set in when $N$ exceeded the optimum. For example, the 4-layer NN with 15 neurons per layer had an average/(minimum-maximum) test set error of 1662.90/(1.07-4114.96) cm$^{-1}$ (indicating a severe local minima problem), and with 25 neurons it was 73.40/(21.55-136.03) cm$^{-1}$. A single hidden layer NN with 75 neurons had an average/maximum test set error of 5.38/(1.17-18.14) cm$^{-1}$. Local minima are particularly severe in multilayer NNs, for example, a rerun of NN training with the same 5 runs for each architecture returned, for a 4-layer NN with 10 neurons per layer, average/(minimum-maximum) test set rmse values of 817.9/(0.38-4082.65) cm$^{-1}$. Singe hidden layer NNs, on the contrary, did not have this problem and returned consistent results.

Opt-GPRNN thus outperformed the conventional NN while avoiding the need to optimize the numbers of layers and neurons and to incur large numbers of epochs with dangers of bad local minima. Even with very suboptimal hyperparameters, opt-GPRNN is competitive with a well-optimized NN. As in GPRNN, each component function is univariate, it is not difficult to find near-optimal hyperparameters.

### 3.2 *Zero-point vibrational energy*

In Figure 3, we show training and test rmse values for different $N$, with and without $W$ optimization, when machine learning the ZPE of 16-atom molecules. We also show the learning curve for the case $N = 100$ (when the test set error levels off) and the corresponding parity plot in Figure 4 (we did not show the parity plot in the previous example as it is visually fully diagonal due to the high accuracy). In this application, it is meaningful to explore dimensionality reduction. The dimensionality reduction regime corresponds to negative abscissa values (where $N<D$). We see a sharp drop and inflection point of the error when $N = D$, indicating that not just formal but the intrinsic dimensionality of the feature space is 16. After $N$ crosses $D$ (at which point the test set error is 1.01 mHa), there is only mild improvement in the test error that levels at 0.96 mHa past $N = 70$. There is no overfitting with increased $N$. Optimization mostly affects results at low $N$, as expected.

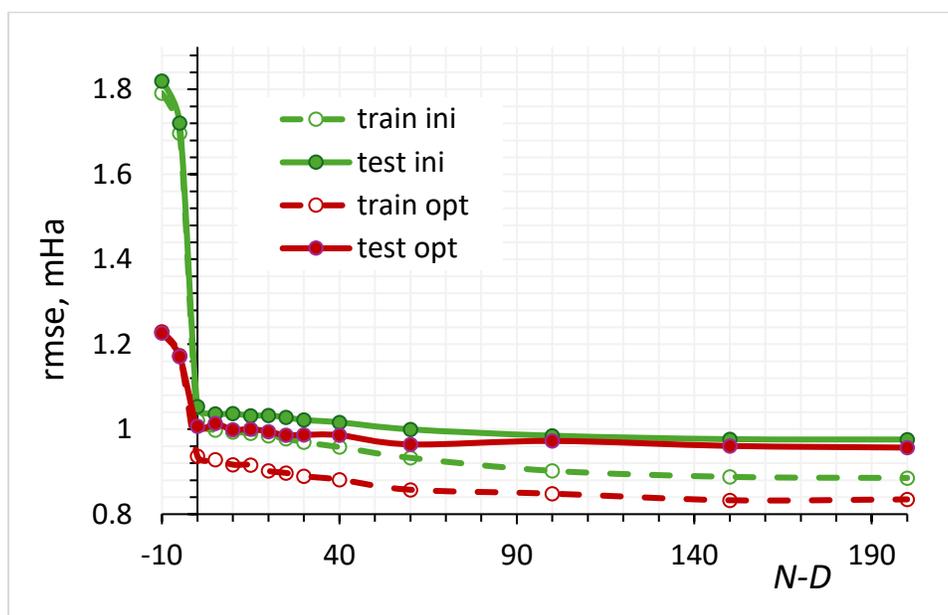

Figure 3. The behavior of training and test set errors when fitting the ZPE with GPRNN ("ini") and opt-GPRNN ("opt") as a function of the number of terms. The abscissa shows the number of redundant coordinates $N - D$.

We also compared the results with conventional NN with optimized numbers of layers and neurons per layer. The best NN was single hidden layer NN with 15 neurons (corresponding to the inflection point of Figure 3) which we obtained average/(minimum-maximum) training set rmse values of 0.84/(0.82-0.86) mHa and average/(minimum-maximum) test set rmse values of 1.01/(1.00-1.01) mHa. Overfitting appeared as the number of neurons was increased beyond optimal, for example, with *N* = 100, average/(minimum-maximum) test set rmse was 1.97/(1.91-2.04) mHa. The learning curve and parity plots for the optimal NN size are shown in Figure 5. In this example, the level of accuracy exemplified by the shape of the parity plots, is typical of materials informatics where achievable accuracy is limited by the predictive power of the features and often by the available dataset sizes. We chose for this example a relatively large dataset so that we could use a large test set, but the extent of learnability of the property is limited, as is common in the field of materials informatics, where featurization is an active topic of research [58–65]. It is for this reason (of limited learnability) that the training stops after a relatively small number of cycles (as opposed to more than 1000 cycles in a high-learnability, high-accuracy regime of the previous example). Nevertheless, it is clear that also in this example opt-GPR-NN is competitive with the best NN.

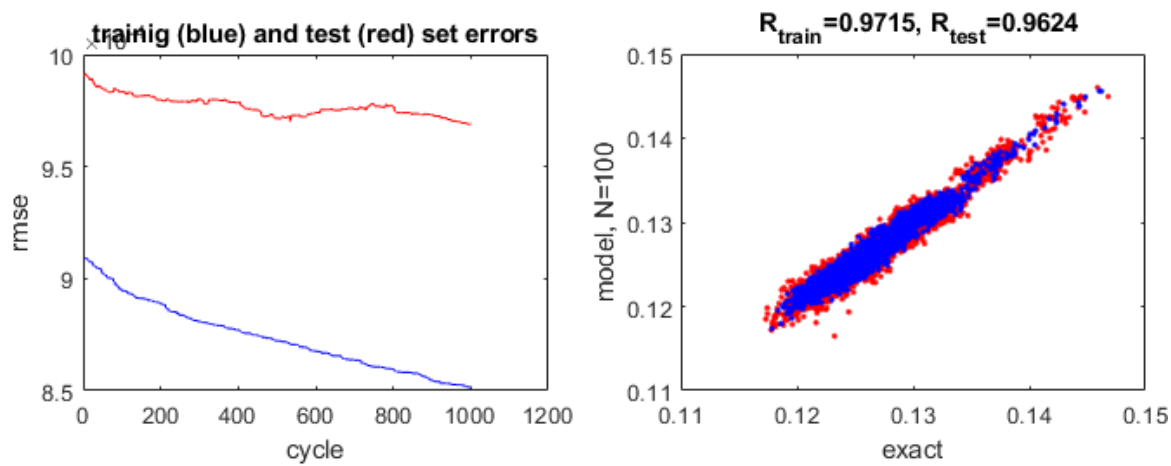

Figure 4. Left: learning curve when machine learning the ZPE with opt-GPRNN with *N* = 100. Right: parity plot. The units for "exact" (ground truth) and "model" (opt-GPRNN) are Hartree. Training data are in blue and test set data are in red in both plots. Corresponding correlation coefficients are also shown.

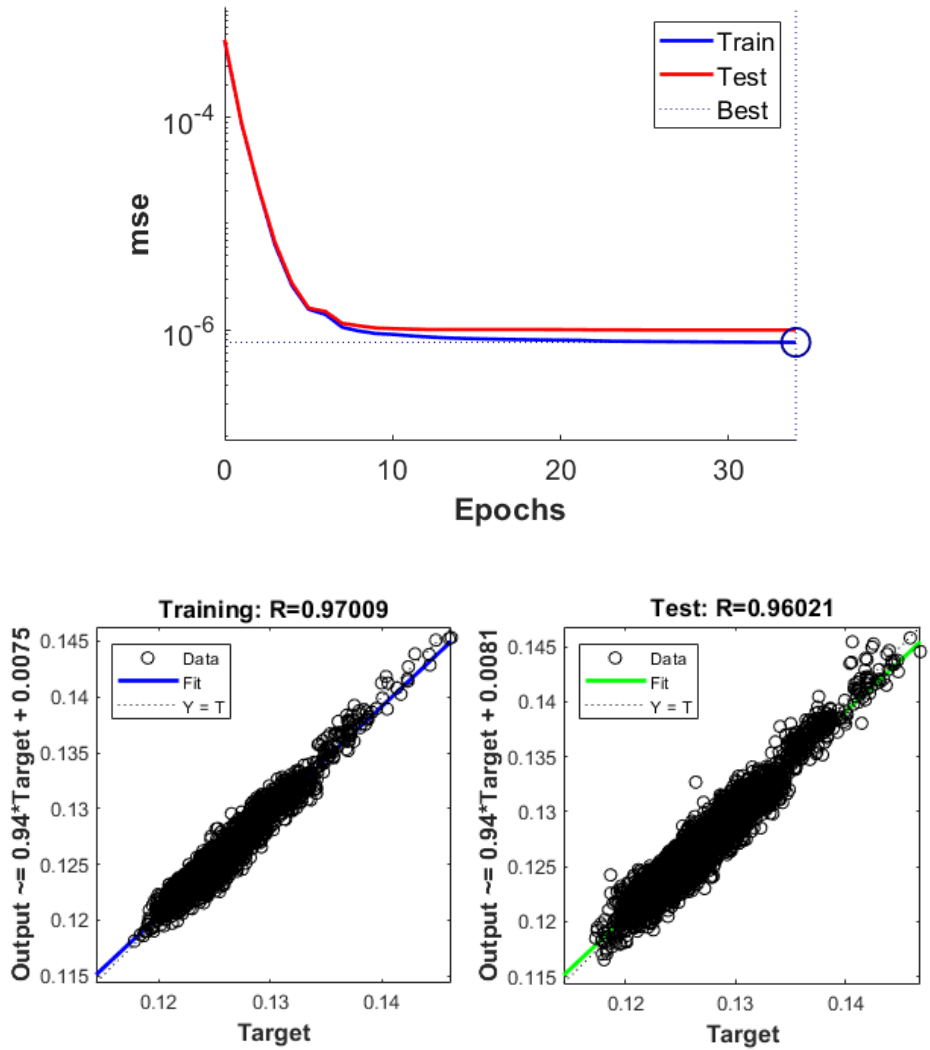

Figure 5. Learning curve (top) and parity plots for training (bottom left) and test (bottom right) sets when machine learning the ZPE with a conventional NN.

## 4 Conclusions

We explored a modification of the GPRNN algorithms with the redundant coordinates optimized by a Monte Carlo algorithm, opt-GPRNN. Optimization leads to much lower errors when such improvement is possible, i.e. when the learnability of the target from the data allows it and when the number of terms $N$ is small (as the best possible error can be obtained with GPRNN with a sufficiently high number of terms even without optimization, the advantage of optimization naturally dwindles at large *N*). The optimal number of terms at which the test set error levels off is smaller due to optimization. While GPRNN was equivalent to a single hidden layer NN with optimal shapes of neuron activation functions and fixed weights, opt-GPRNN is equivalent to a single hidden layer NN with optimal shapes of neuron activation functions and optimal neuron arguments, increasing expressive power. While $\theta_i(\boldsymbol{W}|\boldsymbol{x})$ of the deep NN of Eq. (3) are fixed-shape activation functions whose arguments are complex non-linear functions of $\boldsymbol{x}$, the fact that $f_n(y_n)$ can have arbitrary shapes partially compensates for the fact that the neuron argument $y_n = \boldsymbol{W}\boldsymbol{x}$ is restricted to a linear form in $\boldsymbol{x}$. The algorithm thus possesses an expressive power closer to that of a multilayer NN and could obviate the need for deep NNs in some applications.

Opt-GPRNN retains a key advantage of GPRNN, namely, stability with respect to overfitting as the number of terms is grown beyond optimal. In our tests, opt-GPRNN outperformed conventional NNs while avoiding the need to optimize the numbers of layers and neurons and to incur dangers of bad local minima. Even with very suboptimal hyperparameters, opt-GPRNN was competitive with a well-optimized NN. As in GPRNN, each component function is univariate, it is not difficult to find near-optimal hyperparameters.

Algorithmically, opt-GPRNN is based on 1$^{\text{st}}$ order additive GPR and Monte Carlo optimization. While this work focuses on the concept and not on computational efficiency, it is worth noting that the only extra cost vs. standard GPR is the summation in the additive kernel, which is perfectly parallelizable. Multiple displacements in MC optimization can also be computed in parallel to accelerate updates. The method thus has some intrinsic parallelizability advantages. We thus hope that opt-GPRNN becomes a useful addition to the machine learning toolbox and inspires further methodologic developments.

## 5  Acknowledgements

We thank Digital Research Alliance of Canada on whose servers some of the calculations were performed. We thank Yemin Thant for helping compile the ZPE dataset.

## 6  Data Availability

The data and codes used in this work can be found at https://github.com/sergeimanzhos/optGPRNN/ .